\documentclass[11pt]{article}

\usepackage{acl_2026}
\usepackage{times}
\usepackage{latexsym}

\usepackage[T1]{fontenc}

\usepackage[utf8]{inputenc}

\usepackage{microtype}

\usepackage{inconsolata}

\usepackage{graphicx}

\usepackage{amsmath}
\usepackage{mathtools}
\usepackage{amsthm}
\usepackage{amsfonts}       
\usepackage{amssymb}

\usepackage[utf8]{inputenc} 
\usepackage[T1]{fontenc}    

\usepackage{url}            
\usepackage{nicefrac}       
\usepackage{xcolor}         
\usepackage{enumitem}
\usepackage{xspace}
\usepackage{etoolbox}

\usepackage{titletoc}
\usepackage{multicol}
\usepackage{multirow}
\usepackage{colortbl}
\usepackage{tabularx}
\usepackage{makecell}
\usepackage{caption}
\usepackage{booktabs}

\usepackage{wrapfig}
\usepackage{rotating}

\usepackage{pifont}
\usepackage{bbding}

\usepackage{listings}
\usepackage{subcaption}
\usepackage{float}
\usepackage[most]{tcolorbox}
\usepackage{algorithm}
\usepackage{algpseudocode}
\usepackage{tikz}
\usetikzlibrary{tikzmark}

\usepackage{amsmath,amsfonts,bm}

\def\eqref#1{equation~\ref{#1}}
\def\Eqref#1{Equation~\ref{#1}}

\def\1{\bm{1}}

\def\vv{{\bm{v}}}

\DeclareMathAlphabet{\mathsfit}{\encodingdefault}{\sfdefault}{m}{sl}
\SetMathAlphabet{\mathsfit}{bold}{\encodingdefault}{\sfdefault}{bx}{n}

\def\gD{{\mathcal{D}}}

\def\gJ{{\mathcal{J}}}

\def\gP{{\mathcal{P}}}

\newcommand{\E}{\mathbb{E}}

\usepackage{amsthm}
\definecolor{myred}{rgb}{0.7, 0.3, 0.0}
\definecolor{myblue}{rgb}{0.2, 0.3, 0.6}
\newcommand{\modelname}{JudgePanel\xspace}
\newcommand{\Sref}[1]{Section~\ref{#1}}

\newcommand{\Fref}[1]{Figure~\ref{#1}}

\newcommand{\Aref}[1]{Appendix~\ref{#1}}

\theoremstyle{remark}

\definecolor{Gray}{gray}{0.95}
\title{\modelname: A Compact Judge with Panel Deliberation via Adaptive Multi-Reward Reinforcement Learning}

\author{
      Yiyue Qian\textsuperscript{\rm }\thanks{\ \ Equal contribution.},
      Shinan Zhang\textsuperscript{\rm }\footnotemark[\value{footnote}],
     Huan Song\textsuperscript{\rm},
      Hannah Marlowe\textsuperscript{\rm}
      \\
      \texttt{\{iamyiyue, shinanz, huanso, marloweh\}@amazon.com} \\
      \\
      \textsuperscript{\rm } AWS Generative AI Innovation Center, USA\\
  }
\begin{document}
\maketitle
\begin{abstract}
The LLM-as-a-Judge paradigm has emerged as a scalable alternative to human evaluation.
However, single-model judges are limited by their inherent model biases, while multi-agent evaluation protocols that mitigate this through diverse deliberation are prohibitively expensive at inference time.
To this end, we propose \textbf{\modelname}, which equips a compact \underline{Judge} model with multi-agent \underline{Panel} deliberation capability. Specifically,
we first train on panel deliberation traces from an ensemble of strong evaluators, capturing structured patterns of discussion, disagreement, and resolution.
To further improve judgment quality beyond SFT, we introduce \textit{AdaReward}, an adaptive multi-reward RL algorithm that dynamically rebalances reward component weights as different objectives saturate at different rates during RL training.
For practical deployment, we further design a lightweight domain specialization module for rapid adaptation to new evaluation domains with few hundred labeled samples.
As a result,
(i) \textit{Novel}: the first framework to equip a single compact judge with multi-agent panel deliberation capability at single-model inference cost;
(ii) \textit{Effective \& Reliable}: JudgePanel with a 14B backbone outperforms judge-specialized models up to 70B across four evaluation benchmarks, demonstrates strong position consistency, and rapidly specializes to new domains with few hundred samples.
\end{abstract}

\section{Introduction}

\begin{figure}[t]
    \centering
    \includegraphics[width=0.9\linewidth]{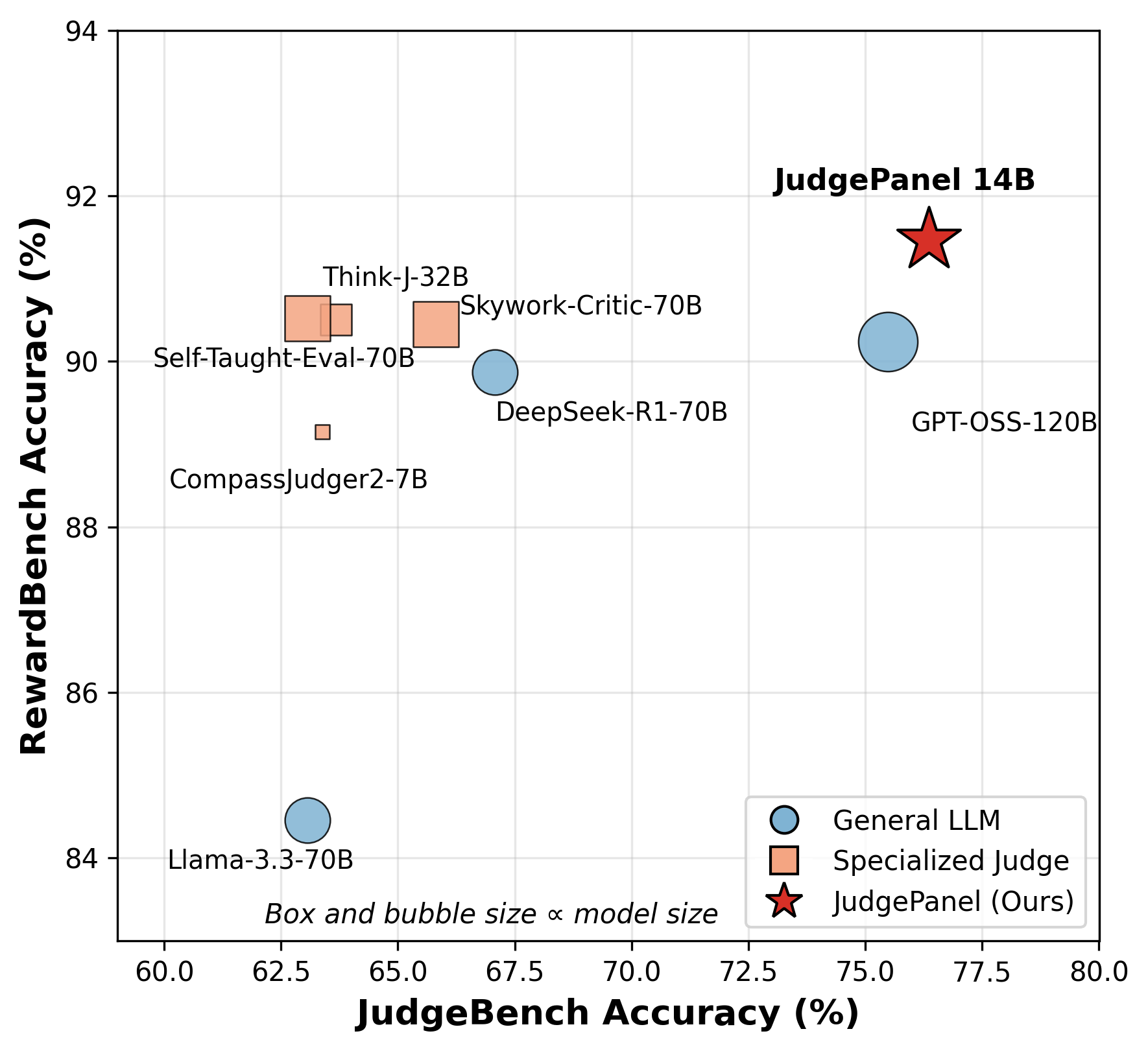}
    \caption{JudgeBench vs.\ RewardBench accuracy. Existing models show inconsistent cross-dataset performance, while JudgePanel achieves the best on both.}
    \label{fig:domain_generalization}
\end{figure}

As large language models (LLMs) continue to advance, reliably evaluating their outputs has become increasingly challenging due to the high cost and limited scalability of human evaluation~\cite{clark2021all, karpinska2021perils,qian2025enhancing}.
The LLM-as-a-Judge paradigm, where LLMs generate chain-of-thought reasoning before producing a verdict, has emerged as a promising alternative~\cite{zheng2023judging, kim2024prometheus, saha2024bsm}.
However, single-model judges are inherently limited by their model biases stemming from training data and optimization, leading to systematic evaluation biases and inconsistent judgments across domains~\cite{chen2024humans, wang2024llmjudgebias}.

To mitigate these biases, recent work employs multi-agent evaluation protocols where multiple LLMs assess, discuss disagreements, and converge on a consensus~\cite{qian2026collabeval, chan2024chateval, saha2024bsm}.
While effective at reducing individual model biases through diverse perspectives, such protocols require deploying multiple large models with multi-round inference, making them prohibitively expensive for large-scale deployment.
This raises a natural question: \textit{can we train a single compact model that reasons like a panel of experts?}

To this end, we propose \textbf{\modelname}, which equips a compact \underline{Judge} model with multi-agent \underline{Panel} deliberation capability.
We first construct panel deliberation traces, capturing structured patterns of discussion, disagreement, and resolution.
By training a compact model on these traces via SFT, we equip it with panel deliberation at single-model inference cost.
However, SFT relies on pattern imitation rather than deep reasoning~\cite{chen2025judgelrm}, limiting the model to reproducing the quality of its training traces without the ability to generalize beyond them.
To address this, we apply multi-reward RL that jointly optimizes judgment quality from multiple perspectives, such as format compliance, discussion quality, and outcome correctness.
Yet in multi-reward training, different components learn at different rates: format compliance saturates early while outcome correctness requires sustained optimization, and fixed weights cannot capture this dynamic.
We therefore propose \textbf{AdaReward}, an adaptive weight learner that dynamically rebalances reward components during RL training, creating a curriculum-like effect without manual tuning.
Furthermore, as illustrated in \Fref{fig:domain_generalization}, even well-trained judges show inconsistent performance across evaluation domains. Specialized judge models excel on some benchmarks while underperforming on others.
For practical deployment, we design a lightweight domain specialization module that rapidly adapts the judge to new domains with a few hundred labeled samples.

We summarize our contributions as follows:
\vspace{-2mm}
\begin{itemize}[leftmargin=*, noitemsep]
    \item \textbf{Novel Framework}: We propose \modelname, the first to equip a single compact judge with panel deliberation capability via adaptive multi-reward RL, achieving multi-agent reasoning quality at single-model inference cost.
    \item \textbf{AdaReward}: We introduce an adaptive multi-reward RL algorithm that dynamically rebalances reward component weights during RL training, shifting focus from saturated components to those still providing useful signals.
    \item \textbf{State-of-the-Art Performance}: \modelname with a 14B backbone outperforms larger judge-specialized models (i.e., 70B) across four evaluation benchmarks, demonstrates strong position consistency, and rapidly specializes to new domains with few hundred samples.
\end{itemize}

\begin{figure*}[t]
    \centering
    \includegraphics[width=1.15\linewidth]{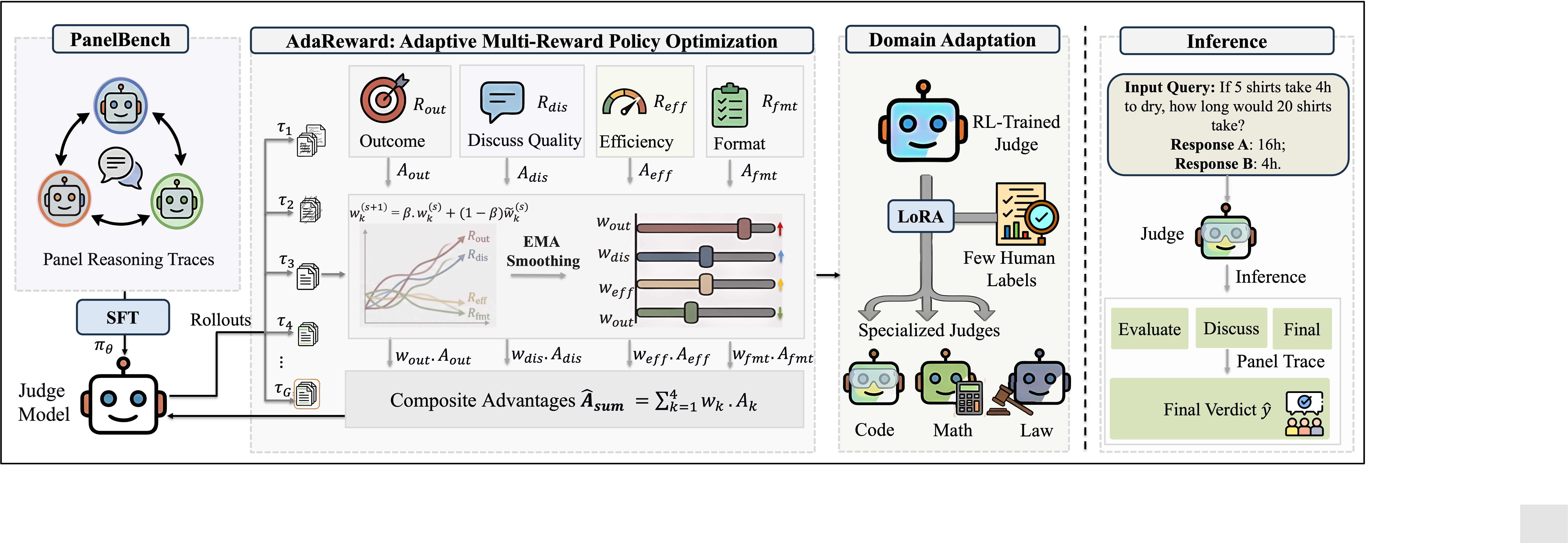}
    \vspace{-0.5in} 
    \caption{The overall framework of \modelname: (a) \textit{PanelBench}: A panel evaluation through multi-round discussion to produce panel deliberation traces for training a judge model $\pi_\theta$ via SFT. (b) \textit{AdaReward}: The judge generates rollouts scored by four reward components. An EMA-based adaptive weight learner dynamically rebalances reward weights, producing composite advantages for policy updates. (c) \textit{Domain Adaptation}: The RL-trained judge is specialized to new domains via LoRA fine-tuning on a few hundred human-labeled samples. (d) \textit{Inference}: At deployment, the compact judge generates a full panel deliberation trace at single-model inference cost.}
    \label{fig: framework}
\end{figure*}
\section{Preliminary}

\noindent\textbf{LLM-as-a-Judge.}
Given an evaluation input $x_i$ (e.g., an instruction with candidate responses), an LLM-as-a-Judge $\pi_\theta$ generates a reasoning trace $t_i$ followed by a verdict $\hat{y}_i$.
During RL training, we sample $G$ rollouts $\{\tau_{i,j}\}_{j=1}^{G}$ from the reference policy $\pi_{\text{ref}}$, where each rollout $\tau_{i,j} = (t_{i,j}, \hat{y}_{i,j})$ contains a reasoning trace and predicted verdict.

\noindent\textbf{Group reward-Decoupled Normalization Policy Optimization (GDPO).}\label{sec:prelim}
To optimize judge with $K$ reward functions $\{R_k\}_{k=1}^{K}$, inspired by GDPO~\cite{liu2026gdpo}, which resolves the \textit{reward signal collapse} of standard Group Relative Policy Optimization (GRPO)~\cite{shao2024deepseekmath} by normalizing each reward before aggregation, we employ GDPO as our base RL algorithm. 
GDPO computes per-reward advantages $A_k^{(i,j)} = ({R_k(\tau_{i,j}) - \text{mean}_j\{R_k\}})/{\text{std}_j\{R_k\}}$ and aggregates them as:\vspace{-0.1in}
\vspace{-1mm}
\begin{align}\label{eq:gdpo_agg}
    A_{\text{sum}}^{(i,j)} &= \sum_{k=1}^{K} w_k \cdot A_k^{(i,j)}, \nonumber \\[-1mm]
    \hat{A}_{\text{sum}}^{(i,j)} &= \frac{A_{\text{sum}}^{(i,j)} - \text{mean}_{\text{batch}}\{A_{\text{sum}}\}}{\text{std}_{\text{batch}}\{A_{\text{sum}}\}},
\vspace{-0.02in}
\end{align}
with fixed weights $\{w_k\}$.
The policy is optimized via a clipped surrogate objective $\gJ_{\text{GDPO}}(\theta)$.

\section{\modelname: Panel Deliberation via Adaptive Multi-Reward RL}
As illustrated in \Fref{fig: framework}, \modelname is a three-stage framework:
(i) \textit{Panel Deliberation}, which generates collaborative evaluation traces for model training;
(ii) \textit{AdaReward}, an adaptive reward weight learner for multi-reward RL training; and
(iii) \textit{Lightweight Domain Specialization}, which specializes the judge to new domains with hundred samples via LoRA training.
\subsection{Panel Deliberation via Multi-Agent Collaboration}\label{sec:data}
\vspace{-1mm}
Existing LLM-as-a-Judge approaches~\cite{whitehouse2026j1, huang2026thinkj, saha2025learning, zhang2025compassjudger2} rely on a single model to generate synthetic training data, creating a reasoning diversity bottleneck that misses the multi-model collaborative deliberation characteristic of high-quality human evaluation.
To overcome this, following the similar mechanism of CollabEval~\cite{qian2026collabeval}, we obtain the reasoning traces samples from Chatbot Arena dataset~\cite{chiang2024chatbot}, a large-scale collection of real user conversations with deployed LLMs.
\begin{figure}[H]
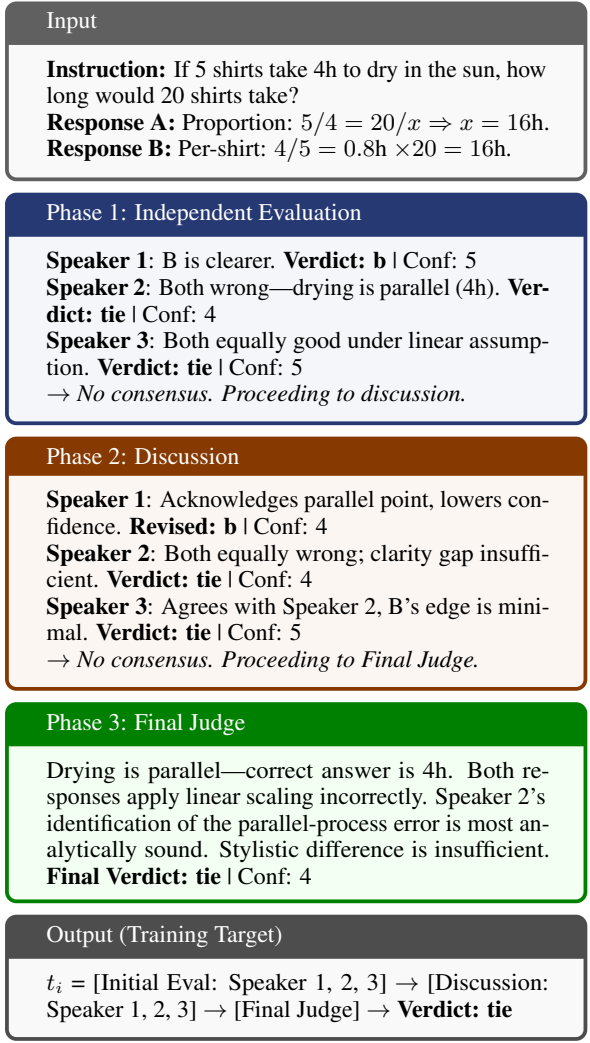

        \vspace{-2mm}
    \centering
    \small
    \begin{tcolorbox}[colback=gray!5!white,colframe=gray!75!black, title={\small Input}, boxsep=1mm, top=1mm, bottom=1mm]
        \footnotesize
        \textbf{Instruction:} If 5 shirts take 4h to dry in the sun, how long would 20 shirts take? \\
        \textbf{Response A:} Proportion: $5/4 = 20/x \Rightarrow x = 16$h. \quad
        \textbf{Response B:} Per-shirt: $4/5 = 0.8$h $\times 20 = 16$h.
    \end{tcolorbox}
    \vspace{-2mm}
    \begin{tcolorbox}[colback=myblue!5!white,colframe=myblue!75!black, title={\small Phase 1: Independent Evaluation}, boxsep=1mm, top=1mm, bottom=1mm]
        \footnotesize
        \textbf{Speaker 1}: B is clearer. \textbf{Verdict: b} | Conf: 5 \\
        \textbf{Speaker 2}: Both wrong---drying is parallel (4h). \textbf{Verdict: tie} | Conf: 4 \\
        \textbf{Speaker 3}: Both equally good under linear assumption. \textbf{Verdict: tie} | Conf: 5 \\
        \textit{$\rightarrow$ No consensus. Proceeding to discussion.}
    \end{tcolorbox}
    \vspace{-2mm}
    \begin{tcolorbox}[colback=myred!5!white,colframe=myred!75!black, title={\small Phase 2: Discussion}, boxsep=1mm, top=1mm, bottom=1mm]
        \footnotesize
        \textbf{Speaker 1}: Acknowledges parallel point, lowers confidence. \textbf{Revised: b} | Conf: 4 \\
        \textbf{Speaker 2}: Both equally wrong; clarity gap insufficient. \textbf{Verdict: tie} | Conf: 4 \\
        \textbf{Speaker 3}: Agrees with Speaker 2, B's edge is minimal. \textbf{Verdict: tie} | Conf: 5 \\
        \textit{$\rightarrow$ No consensus. Proceeding to Final Judge.}
    \end{tcolorbox}
    \vspace{-2mm}
    \begin{tcolorbox}[colback=green!5!white,colframe=green!50!black, title={\small Phase 3: Final Judge}, boxsep=1mm, top=1mm, bottom=1mm]
        \footnotesize
        Drying is parallel---correct answer is 4h. Both responses apply linear scaling incorrectly. Speaker 2's identification of the parallel-process error is most analytically sound. Stylistic difference is insufficient. \textbf{Final Verdict: tie} | Conf: 4
    \end{tcolorbox}
    \vspace{-2mm}
    \begin{tcolorbox}[colback=gray!5!white,colframe=gray!60!black, title={\small Output (Training Target)}, boxsep=1mm, top=1mm, bottom=1mm]
        \footnotesize
        $t_i$ = [Initial Eval: Speaker 1, 2, 3] $\rightarrow$ [Discussion: Speaker 1, 2, 3] $\rightarrow$ [Final Judge] $\rightarrow$ \textbf{Verdict: tie}
    \end{tcolorbox}
    \vspace{-2mm}
    \caption{Simplified three-stage panel reasoning trace. }
    \label{fig:panel_traces}
    \vspace{-0.2in}
\end{figure}

As illustrated in \Fref{fig:panel_traces}, the panel follows a three-phase protocol:
(i) \textbf{\textit{Independent Evaluation}}: each agent independently produces a structured evaluation $e_i^{(m)}$, verdict $v_i^{(m)}$, and confidence $c_i^{(m)}$;
(ii) \textbf{\textit{Multi-Round Discussion}}: triggered when no consensus is reached, evaluators review the full history and provide updated assessments with explicit agreements, disagreements, and revised reasoning; and
(iii) \textbf{\textit{Final Judgment}}: when discussion fails to converge or reaches $d_{\max}$ rounds, a final judge $\pi^{(\text{final})}$ analyzes the complete history and produces the verdict $\hat{y}_i$.
We retain only samples where the panel verdict matches the ground-truth label $y_i$, yielding ${\sim}$8,700 correct panel traces for training.
By training on these collaborative traces, the compact judge model (i.e., 14B) acquires panel deliberation capabilities, which learns to generate independent evaluation, disagreement identification, deliberation, and consensus.
\subsection{AdaReward: Adaptive Multi-Reward Policy Optimization}\label{sec:adgdpo}

As established in Section \ref{sec:prelim}, GDPO resolves the reward signal collapse problem inherent in GRPO's multi-reward optimization by decoupling per-reward normalization.
However, GDPO combines the normalized advantages with fixed weights $\{w_k\}$ in Equation \ref{eq:gdpo_agg} that remain constant throughout training.
In judge training, we observe that reward components exhibit vastly different convergence dynamics.
For instance, format compliance saturates quickly at around 50 steps, while outcome correctness, discussion quality, and efficiency are learning throughout training.
To this end, we propose \textbf{AdaReward}, an adaptive weight learner that dynamically learns reward component weights based on their training progress. 

\subsubsection{Multi-Aspect Reward Design}\label{sec:reward_design}

We design $K=4$ reward functions that jointly capture both judgment correctness and reasoning quality.
Each reward $R_k(\tau_{i,j})$ evaluates an aspect of the panel deliberation trace $\tau_{i,j}$. The detailed scoring of each reward is described in \Aref{appendix:reward_details}.

\noindent\textbf{Fine-Grained Outcome Reward} $R_{\text{out}}$.
Rather than a binary correct/incorrect reward, we design a graded outcome reward that diagnoses \textit{where the reasoning went wrong}: whether the error originated in the initial evaluation, was introduced during discussion, or occurred at the final judgment stage. It provides richer signal than binary rewards.

\noindent\textbf{Discussion Quality Reward} $R_{\text{dis}}$.
This reward measures whether the multi-round discussion produces genuine deliberation and whether speakers actually revise their positions based on others' arguments, rather than merely restating their evaluations.
Crucially, this reward evaluates the quality of the deliberation process itself, rewarding genuine position changes that lead to correct outcomes rather than cosmetic flips.

\noindent\textbf{Efficiency Reward} $R_{\text{eff}}$.
This reward evaluates whether the model's decision to engage in or skip discussion was beneficial, rewarding efficient consensus and penalizing harmful deliberation. $R_{\text{eff}}$ explicitly conditions on correctness: the strongest penalty targets cases where discussion flipped a correct initial majority to a wrong final verdict, directly discouraging counterproductive deliberation.

\noindent\textbf{Format Compliance Reward} $R_{\text{fmt}}$.
This reward ensures the generated trace follows the multi-agent collaboration structure (required section tags, $M$ speakers with complete evaluation fields, proper discussion structure), which is essential for parsing and calculating the other rewards correctly.

\subsubsection{Adaptive Weighting via EMA Learner}\label{sec:ema}

With the four reward components defined, we now address the core limitation of fixed-weight aggregation in existing RL methods.
In GDPO (\Eqref{eq:gdpo_agg}), the composite advantage is $A_{\text{sum}}^{(i,j)} = \sum_{k=1}^{K} w_k \cdot A_k^{(i,j)}$ with constant $w_k$.
When a component $k$ saturates (i.e., $V_k \triangleq \E[(A_k^{(i,j)})^2] \approx 0$), all rollouts receive nearly identical rewards for that component, providing no useful learning signal after GDPO's per-reward normalization.
With fixed weights, when a component saturates, (e.g., $R_{\text{dis}}$), its contribution to the composite advantage vanishes, but its weight cannot be redistributed to harder components (e.g., $R_{\text{out}}$ and $R_{\text{eff}}$) that still provide learning signals.

\noindent\textbf{EMA-Based Adaptive Weight Learner.}
Our key idea is to adjust each weight proportionally to how far its reward is from a target score, pushing harder on lagging components and easing off on saturated ones.
Specifically, we define an gap signal $\delta_k^{(s)} = \rho_k - \bar{R}_k^{(s)}$ measuring the gap between the target $\rho_k$ and the current batch-average reward $\bar{R}_k^{(s)}$ for component $k$ at step $s$.
The weight update uses \textit{asymmetric scaling}: when a component falls behind its target ($\delta_k > 0$), its weight increases aggressively; when it exceeds the target ($\delta_k \leq 0$), the weight decreases conservatively. Concretely, we first compute a target weight $\tilde{w}_k^{(s)}$:

\begin{equation}\label{eq:target_weight}
\resizebox{0.95\linewidth}{!}{$
    \tilde{w}_k^{(s)} = w_k^{(s)} \cdot \begin{cases}
        1 + \alpha^+ \cdot \delta_k^{(s)} & \text{if } \delta_k^{(s)} > 0 \;\; \text{(below target)} \\
        1 + \alpha^- \cdot \delta_k^{(s)} & \text{if } \delta_k^{(s)} \leq 0 \;\; \text{(above target)},
    \end{cases}
$}
\end{equation}
\vspace{-1mm}
where $\alpha^+ > \alpha^-$ ensures faster recovery for underperforming components than decay for well-performing ones.
The target weight is then clipped to $[w_k^{\min}, w_k^{\max}]$ before updates (Eq.~\ref{eq:ema}), ensuring no single component dominates or vanishes regardless of the gap magnitude.
The final weight is updated via exponential moving average (EMA):
\begin{equation}\label{eq:ema}
    \resizebox{0.7\linewidth}{!}{$
    w_k^{(s+1)} = \beta \cdot w_k^{(s)} + (1 - \beta) \cdot \tilde{w}_k^{(s)},
$}
\end{equation}
\vspace{-1mm}
where $\beta \in (0, 1)$ controls the adaptation speed (lower $\beta$ means faster response to new targets).
Weights are updated every $\Delta$ steps, balancing responsiveness (frequent updates) against stability (allowing the policy to learn from current weights before rebalancing).

\noindent\textbf{Intuition.}
Fixed weights apply constant correction regardless of how far each reward is from its target, while AdaReward's correction magnitude scales with the gap $\delta_k$, pushing harder when far from target and easing off when close.
This creates a curriculum-like effect: as easy components (e.g., $R_{\text{fmt}}$) reach their targets, the learner reallocates weight to harder components (e.g., $R_{\text{out}}$) where $\delta_k$ remains large.

Furthermore, Table~\ref{tab:analysis}(f) shows that adaptive weights outperforms fixed weights.

\noindent\textbf{Full AdaReward Objective.}
Combining the decoupled normalization with the adaptive weights, the AdaReward objective is formulated as:

\vspace{-4mm}
\begin{align}\label{eq:adgdpo}
    &\gJ_{\text{Ada}}(\theta) = \E_{\substack{x_i \sim \gD,\, \{\tau_{i,j}\}_{j=1}^{G} \sim \pi_{\text{ref}}}} \Bigg[ \frac{1}{G} \sum_{j=1}^{G} \frac{1}{|\tau_{i,j}|} \sum_{l=1}^{|\tau_{i,j}|} \nonumber \\[-2mm]
    &\min\!\big( r_{l}(\theta)\, \hat{A}_{\text{sum}}^{(i,j,s)},\, \text{clip}(r_{l}(\theta), 1{-}\varepsilon, 1{+}\varepsilon)\, \hat{A}_{\text{sum}}^{(i,j,s)} \big) \Bigg],
\end{align}
\vspace{-2mm}
where $r_l(\theta) = \frac{\pi_\theta(\tau_{i,j}^l \mid x_i, \tau_{i,j}^{<l})}{\pi_{\text{ref}}(\tau_{i,j}^l \mid x_i, \tau_{i,j}^{<l})}$ is the importance sampling ratio at token position $l$.
$\hat{A}_{\text{sum}}^{(i,j,s)}$ is the batch-normalized composite advantage computed using the time-varying weights $\{w_k^{(s)}\}$ at step $s$.

\subsection{Lightweight Domain Specialization}\label{sec:da}
As shown in Figure~\ref{fig:domain_generalization}, even well-trained judges exhibit inconsistent performance across different evaluation domains.
These judge-specialized models trained on narrow distributions struggle to generalize. For instance, Skywork-Critic-70B~\cite{liu2026skywork} performs well on RewardBench but poorly on JudgeBench.
In real-world deployment, different use cases demand different domain expertise, yet existing approaches do not provide a mechanism for rapid specialization.
This motivates a lightweight mechanism for rapid domain specialization without retraining from scratch.

Our key insight is that a practitioner can quickly curate a small set of human-labeled samples (e.g., 200--500) for a new evaluation domain, then apply the panel protocol (\Sref{sec:data}) to automatically generate domain-specific panel deliberation traces from these labels.

We further fine-tune the RL-trained judge using LoRA~\cite{hu2021lora} on these traces.
LoRA is chosen over full fine-tuning to preserve general deliberation capabilities acquired during RL while efficiently injecting domain-specific knowledge with minimal parameters.

The full training pipeline of \modelname is summarized in Appendix Algorithm~\ref{alg:pipeline}, and all hyperparameters settings are provided in \Aref{appendix:hyperparameters}.

\section{Experiments}
\vspace{-2mm}
\begin{table*}[t]
      \centering
      \caption{Accuracy performance (\%) comparison among general LLMs and judge-specialized models on JudgeBench, RewardBench, RMBench, and PPE. Avg.\ Rank is the mean
  rank across all four benchmarks (lower is better). The bolded numbers are the best results and the underlined numbers indicate the runner-up results.
      } \label{tab:main_exp}
      \resizebox{1.0\linewidth}{!}
      {
      \begin{tabular}{clccccc}
      \toprule
           \textbf{Type} & \textbf{Model} & \textsc{JudgeBench} & \textsc{RewardBench} & \textsc{RMBench} & \textsc{PPE} & \textsc{Avg. Rank}$\downarrow$ \\
           \midrule
           \multirow{4}{*}{\makecell{General\\LLM}} 
           & Qwen2.5-7B-Instruct & 57.66 & 78.81 & 65.69 & 38.14 & 11.0 \\
           & DeepSeek-R1-8B & 56.56 & 76.25 & 74.97 & 37.89 & 10.2 \\
           & Llama-3.3-70B & 63.06 & 84.46 & 73.55 & 41.67 & 7.6 \\
           & DeepSeek-R1-70B & 67.08 & 89.87 & 85.17 & 42.89 & 3.8 \\
           \midrule
           \multirow{8}{*}{\makecell{Specialized\\Judge}}
           & JudgeLRM-3B\textsuperscript{\tiny\,RL} & 60.00 & 64.09 & 58.58 & 36.00 & 12.8 \\
           & JudgeLRM-7B\textsuperscript{\tiny\,RL} & 55.84 & 76.28 & 66.54 & 37.80 & 12.2 \\
           & CompassJudger2-7B\textsuperscript{\tiny\,SFT+RL} & 63.39 & 89.15 & 75.33 & 40.26 & 6.2 \\
           & ArmoRM-8B\textsuperscript{\tiny\,SFT} & 55.81 & 88.44 & 69.23 & 37.26 & 11.2 \\
           & DeepSeek-GRM-27B\textsuperscript{\tiny\,SFT+RL} & 56.46 & 79.19 & 69.19 & 38.31 & 10.5 \\
           & Think-J-32B\textsuperscript{\tiny\,SFT+RL} & 63.68 & 90.50 & 79.80 & 40.15 & 5.2 \\
           & Self-Taught-Eval-70B\textsuperscript{\tiny\,SFT} & 63.06 & 90.52 & 74.64 & 41.71 & 5.4 \\
           & Skywork-Critic-70B\textsuperscript{\tiny\,SFT} & 65.81 & 90.45 & 74.50 & 40.49 & 5.8 \\
           \midrule
           \multirow{2}{*}{Ours}
           & \cellcolor{Gray} Qwen3-14B (main) & \cellcolor{Gray} \underline{74.45} & \cellcolor{Gray} \underline{91.33} & \cellcolor{Gray} \underline{87.54} &
  \cellcolor{Gray} \underline{45.61} & \cellcolor{Gray} \underline{2.0} \\
           & \cellcolor{Gray} Qwen3-14B (w/ DA) & \cellcolor{Gray} \textbf{77.93} & \cellcolor{Gray} \textbf{93.33} & \cellcolor{Gray} \textbf{89.99} & \cellcolor{Gray}
  \textbf{46.69} & \cellcolor{Gray} \textbf{1.0} \\
      \bottomrule
      \end{tabular}
      }
  \end{table*}
\subsection{Datasets and Baseline Models}\label{sec:datasets}

\textbf{PanelBench.}
Following the panel protocol (\Sref{sec:data}), we construct \textit{PanelBench} from Chatbot Arena~\cite{chiang2024chatbot} (20K pairwise samples with human labels).
After filtering out incorrect verdicts, we retain 8,500 panel deliberation traces for SFT and RL training.
Mention that, PanelBench contains no overlapping input queries or responses with any evaluation benchmark test set or domain adaptation data.

\noindent\textbf{Benchmarks.}
We evaluate on four established judge benchmarks (details in \Aref{appendix:benchmarks}):
(1)~\textit{JudgeBench}~(JB)~\cite{tan2025judgebench}: 620 challenging pairs across knowledge, reasoning, math, and coding;
(2)~\textit{RewardBench}~(RB)~\cite{lambert2025rewardbench}: 2,985 prompt-chosen-rejected trios;
(3)~\textit{RMBench}~(RM)~\cite{liu2025rmbench}: 3,981 samples testing sensitivity to subtle content differences;
(4)~\textit{PPE}~\cite{frick2025evaluate}: 16,038 multilingual pairwise preference samples.

\noindent\textbf{Domain Specialization Data.}
To simulate real-world deployment, we generate DA data from related but independent datasets for each benchmark.
As some DA source datasets only provide ground-truth labels without preference pairs, we use Qwen2.5-72B-Instruct~\cite{yang2025qwen3} to generate incorrect responses, constructing pairwise preference samples.
We then run our panel protocol on these pairs, retaining 200--500 correctly predicted samples per benchmark for LoRA adaptation (details in \Aref{appendix:benchmarks}).

\noindent\textbf{Baselines.}
We compare against (1)~\textit{General LLMs}: Qwen2.5-7B-Instruct~\cite{yang2025qwen3}, DeepSeek-R1-8B~\cite{guo2025deepseekr1}, 
Llama-3.3-70B~\cite{grattafiori2024llama3}, DeepSeek-R1-Distill-Llama-70B~\cite{shao2024deepseekmath};
and
(2)~\textit{Judge-specialized models}: JudgeLRM-3B/7B\textsuperscript{RL}~\cite{chen2025judgelrm}, CompassJudger2-7B\textsuperscript{SFT+RL}~\cite{zhang2025compassjudger2}, ArmoRM-8B\textsuperscript{SFT}~\cite{wang2024interpretable}, DeepSeek-GRM-27B\textsuperscript{SFT+RL}~\cite{yang2024regularizing}, Think-J-32B\textsuperscript{SFT+RL}~\cite{huang2026thinkj}, Self-Taught-Eval-70B\textsuperscript{SFT}~\cite{wang2024selftaught}, Skywork-Critic-70B\textsuperscript{SFT}~\cite{liu2026skywork}.

\noindent\textbf{Our Models.}
We train JudgePanel using Qwen3-14B~\cite{yang2025qwen3} with three stages: SFT on PanelBench, AdaReward RL, and LoRA domain adaptation. For fair comparison, we provide all baseline and our model parsed accuracy as the performance results in this paper. 
The detailed \textbf{hyper-parameters} for each stage are \textbf{in \Aref{appendix:hyperparameters}}.

\subsection{Main Results}\label{sec:main_results}

\begin{table*}[t]
    \centering
    \caption{Position consistency and stable accuracy analysis (selected models). Cons.~(\%,$\uparrow$): percentage of samples where verdicts correctly track swaps. Stable Acc.~(\%,$\uparrow$): percentage of samples correctly predicted in \textit{both} orderings. 
    }\label{tab:position_exp} 
    \resizebox{1.0\linewidth}{!}
    {
    \begin{tabular}{clcccccccc}
    \toprule
         \multirow{2}{*}{\textbf{Type}} & \multirow{2}{*}{\textbf{Model}} & \multicolumn{2}{c}{\textsc{JudgeBench}} & \multicolumn{2}{c}{\textsc{RewardBench}} & \multicolumn{2}{c}{\textsc{RMBench}} & \multicolumn{2}{c}{\textsc{PPE}} \\
         \cmidrule(lr){3-4}\cmidrule(lr){5-6}\cmidrule(lr){7-8}\cmidrule(lr){9-10}
          & & Cons. & Stable Acc. & Cons. & Stable Acc. & Cons. & Stable Acc. & Cons. & Stable Acc. \\
         \midrule
         \multirow{2}{*}{\makecell{General\\LLM}}
         & Llama-3.3-70B & 68.87 & 46.94 & 88.51 & 79.77 & 77.24 & 62.35 & 74.45 & 34.78 \\
         & DeepSeek-R1-70B & 71.95 & 52.97 & 91.73 & 85.86 & 86.82 & 79.05 & 73.57 & 34.66 \\
         \midrule
         \multirow{3}{*}{\makecell{Specialized\\Judge}}
         & CompassJudger2-7B\textsuperscript{\tiny\,SFT+RL} & 75.16 & 49.68 & \underline{93.53} & 88.94 & 78.60 & 65.11 & 72.41 & 34.93 \\
         & Self-Taught-Eval-70B\textsuperscript{\tiny\,SFT} & 70.48 & 47.74 & 89.71 & 85.52 & 76.82 & 62.53 & 74.16 & 33.87 \\
         & Skywork-Critic-70B\textsuperscript{\tiny\,SFT} & 79.35 & 54.68 & 92.73 & 88.94 & 80.83 & 65.54 & 71.43 & 35.13 \\
         \midrule
         \multirow{2}{*}{Ours}
         & \cellcolor{Gray} Qwen3-14B (main) & \cellcolor{Gray} \underline{82.61} & \cellcolor{Gray} \underline{67.98} & \cellcolor{Gray} 92.35 & \cellcolor{Gray} \underline{89.21} & \cellcolor{Gray} \underline{89.82} & \cellcolor{Gray} \underline{89.54} & \cellcolor{Gray} \underline{75.32} & \cellcolor{Gray} \underline{35.45} \\
         & \cellcolor{Gray} Qwen3-14B (w/ DA) & \cellcolor{Gray} \textbf{83.61} & \cellcolor{Gray} \textbf{69.61} & \cellcolor{Gray} \textbf{94.82} & \cellcolor{Gray} \textbf{90.00} & \cellcolor{Gray} \textbf{90.46} & \cellcolor{Gray} \textbf{89.78} & \cellcolor{Gray} \textbf{77.30} & \cellcolor{Gray} \textbf{36.62} \\
    \bottomrule
    \end{tabular}
    }
\end{table*}

\noindent\textbf{Effectiveness.}
Table~\ref{tab:main_exp} reports the accuracy performance on four benchmarks.
We highlight the following key findings:
(i)~Judge-specialized models do not consistently outperform general LLMs; 
Even large judge-specialized models such as Skywork-Critic-70B and Self-Taught-Eval-70B fail to maintain consistently strong performance across all benchmarks, confirming the cross-domain generalization challenge identified in \Fref{fig:domain_generalization}.
(ii)~RL-only judges underperform SFT-based judges at comparable sizes, indicating RL alone without quality SFT initialization is insufficient.
(iii)~JudgePanel with domain adaptation achieves the highest average rank, outperforming all baseline models with much bigger size.
(iv)~Without DA, \modelname already surpasses all baselines and judge-specialized models.
We note that domain adaptation is a general technique applicable to any model; we report both settings for completeness. We would also discuss the effects of domain adaption samples on the model performance later.

\noindent\textbf{Reliability.}
Table~\ref{tab:position_exp} presents position consistency and stable accuracy analysis on selected strong baselines.
We can found out most judge-specialized models suffer from positional bias. 
JudgePanel achieves the highest consistency and stable accuracy on all benchmarks, surpassing all 70B models despite being 5 $\times$ smaller.
This confirms that panel deliberation, which requires evaluating from multiple perspectives before reaching a verdict, encourages content-grounded evaluation that is robust to input ordering.

\begin{table*}[t]
    \centering
     \caption{
        \modelname{} analysis. The highlighted row in gray is our method. RB = RewardBench, RM = RMBench.
    } \label{tab:analysis}
    \vspace{-0.05in}
    \begin{minipage}[t]{0.30\textwidth}
     \centering
       (a) Generated Panel Trace Quality. 
        \resizebox{\linewidth}{!}{  
        \begin{tabular}{lcc}
        \toprule
         \textbf{Metric} & \textbf{RB} & \textbf{RM} \\
           \midrule
           Constructive Change (\%,$\uparrow$) & 91.0 & 85.7 \\
           Jaccard Distance ($\uparrow$) & 0.629 & 0.639 \\
           4-gram Overlap (\%,$\downarrow$) & 14.4 & 15.5 \\
           Avg.\ Rounds & 1.71 & 1.64 \\
        \bottomrule
        \end{tabular}
        } 
    \end{minipage}
    \hfill
    \begin{minipage}[t]{0.33\textwidth}
            \centering
            (b) Multi-Agent Protocol  vs.\ JudgePanel vs.\ Think-J.
            \resizebox{\linewidth}{!}{
            \begin{tabular}{lccc}
            \toprule
              & \textbf{Multi-Agent} & \textbf{JudgePanel} & \textbf{Think-J} \\
              \midrule
              RewardBench & 94.54 & 91.33 & 90.50 \\
              PPE & 46.28 & 45.61 & 40.15 \\
              \midrule
              Parameters & 260B & 14B & 32B \\
              Latency (n=100) & 2.9 min & 1.5 min & 1.0 min \\
            \bottomrule
            \end{tabular}
            } 
    \end{minipage}
    \hfill
    \begin{minipage}[t]{0.31\textwidth}
        \centering
        (c) Label Only vs.\ Single COT vs.\ Voting vs.\ Panel Trace.
        \resizebox{\linewidth}{!}{
        \begin{tabular}{lcc}
        \toprule 
        \textbf{Training Data} & \textbf{RB} & \textbf{RM} \\
        \midrule
        Label Only (SFT) & 87.84 & 76.26 \\
        Single COT (SFT) & 88.14 & 77.64 \\
        Majority Voting (SFT) & \textit{88.50} & \textit{78.56} \\
        \cellcolor{Gray}Panel Trace (Our SFT)  & \cellcolor{Gray} 89.53 & \cellcolor{Gray} 83.77 \\
        \bottomrule
        \end{tabular}
        }
    \end{minipage}
    \\[4mm]
    \begin{minipage}[t]{0.31\textwidth}
            \centering
            (d) Backbone Agnostic.
            \resizebox{\linewidth}{!}{
            \begin{tabular}{lcc}
            \toprule
             \textbf{Model} & \textbf{RB} & \textbf{RM} \\
              \midrule
             Qwen3-32B (base) & 88.21 & 74.52 \\
             \cellcolor{Gray}Qwen3-32B (JudgePanel) & \cellcolor{Gray} 92.23 & \cellcolor{Gray} 89.24 \\
             \midrule
             Qwen3-14B (base) & 86.24 & 72.25 \\
             \cellcolor{Gray}Qwen3-14B (JudgePanel) & \cellcolor{Gray} 91.33 & \cellcolor{Gray} 87.54 \\
            \bottomrule
            \end{tabular}
            } 
    \end{minipage}
    \hfill
    \begin{minipage}[t]{0.33\textwidth}
        \centering 
        (e) Reward Component Ablation.
        \resizebox{\linewidth}{!}{
        \begin{tabular}{lcc}
        \toprule
        \textbf{Reward Setting} & \textbf{RB} & \textbf{RM} \\
        \midrule
        w/o Discussion Quality & 88.24 & 84.74 \\
        w/o Efficiency & 88.64 & 85.16 \\
        w/o Format Compliance & 89.01 & 85.58 \\
        \cellcolor{Gray}Full (ours) & \cellcolor{Gray} 91.33 & \cellcolor{Gray} 87.54 \\

        \bottomrule
        \end{tabular}
        }
    \end{minipage}
    \hfill
    \begin{minipage}[t]{0.32\textwidth}
        \centering
        (f) GDPO vs.\ GRPO.
        \resizebox{\linewidth}{!}{
        \begin{tabular}{lcc}
        \toprule
          \textbf{RL Strategy} & \textbf{RB} & \textbf{RM} \\
          \midrule
          GRPO (fixed weights) & 87.45 & 83.98 \\
          GRPO (adaptive weights) & 89.98 & 85.64 \\
          GDPO (fixed weights) & 89.74 & 84.98 \\
          \cellcolor{Gray}GDPO (adaptive, ours) & \cellcolor{Gray} 91.33 & \cellcolor{Gray} 87.54 \\
        \bottomrule
        \end{tabular}
        }
    \end{minipage}
\end{table*}

\subsection{Ablation and Analysis}\label{sec:ablation}

\begin{figure}[t]
    \centering
    \includegraphics[width=\linewidth]{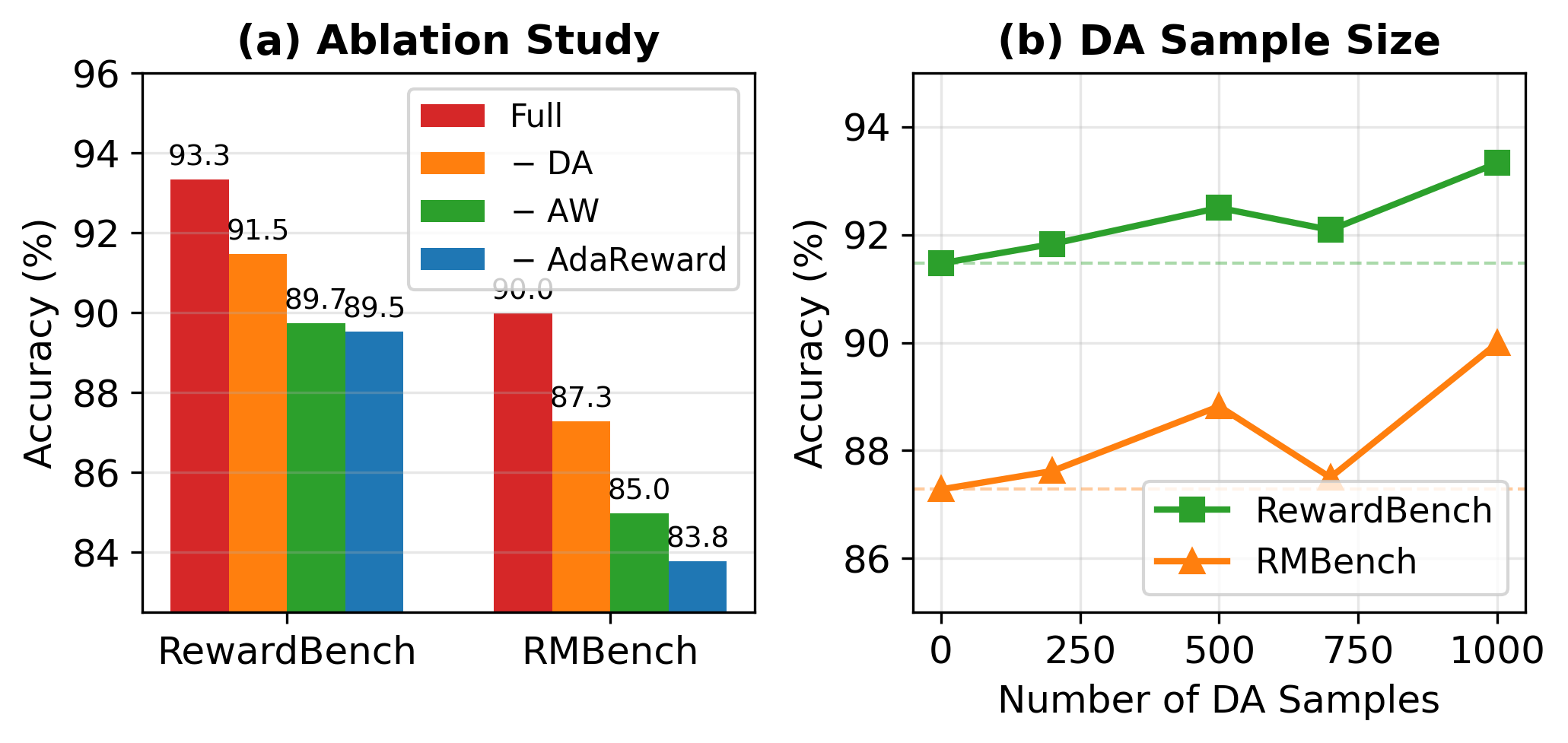}
    \caption{(a) Progressive ablation on RewardBench and RMBench. (b) DA sample size analysis; dashed lines indicate JudgePanel w/o DA.} 
    \label{fig:ablation}
\end{figure}

\textbf{Component Ablation and Reward Ablation Study.}
To validate each component, Figure~\ref{fig:ablation}(a) shows progressive ablation on RewardBench and RMBench.
Removing DA ($-$DA) drops performance by 1.86/2.71, confirming domain specialization is beneficial.
Further replacing adaptive weights with fixed weights ($-$AW) causes an additional 1.73/2.30 drop, and removing RL entirely ($-$AdaReward, SFT only) widens the gap to 3.80/6.22 below the full system.
Table~\ref{tab:analysis}(e) further shows that each reward component is essential: removing Discussion Quality ($-$3.09/$-$2.8) or Efficiency ($-$2.69/$-$2.38) each causes notable degradation, and removing Format Compliance also hurts, confirming that all four rewards capture distinct aspects of panel deliberation quality. To further verify the effectiveness of AdaReward, Table~\ref{tab:analysis}(f) shows our adaptive reward module leads to better final accuracy: GDPO with adaptive weights outperforms both GRPO and GDPO with fixed weights.

\noindent\textbf{Panel Deliberation Analysis.}
To examine whether JudgePanel genuinely deliberates rather than mimicking format, Table~\ref{tab:analysis}(a) measures four metrics on RewardBench and RMBench: Constructive Change Rate (91.0\%/85.7\%) is far above the $\sim$50\% random baseline, showing vote changes lead to correct answers; Jaccard Distance (0.629/0.639) confirms diverse reasoning across speakers; 4-gram Overlap (14.4\%/15.5\%) indicates discussion introduces novel content rather than repeating initial evaluations; and Average Rounds (1.71/1.64) shows the model adaptively engages in deliberation when needed (detailed metric definitions in \Aref{appendix:metrics}).

Beyond quality, we compare against the multi-agent protocol (CollabEval~\cite{qian2026collabeval}) as the quality ceiling and Think-J as a representative COT baseline. Table~\ref{tab:analysis}(b) shows JudgePanel (14B) achieves 91.33/45.61 on RB/PPE, approaching the multi-agent protocol (94.54/46.28) at half the latency (1.5 vs.\ 2.9 min for 100 samples). Think-J-32B achieves lower latency (1.0 min) but with lower accuracy (90.50/40.15), demonstrating that panel deliberation provides a better accuracy-efficiency tradeoff than single-model COT.

To isolate the data contribution, Table~\ref{tab:analysis}(c) compares four SFT strategies on the same Qwen3-14B backbone: Panel Trace SFT outperforms Label Only, Single COT, and Majority Voting (independent evaluations without discussion), demonstrating that multi-round deliberation provides richer supervision than either single-model reasoning or simple multi-perspective aggregation (qualitative examples in \Aref{appendix:case_examples}).

\noindent\textbf{Generalizability.}
To verify JudgePanel is not tied to a specific backbone, Table~\ref{tab:analysis}(d) shows applying it to Qwen3-32B yields +4.02/+14.72 on RB/RM, comparable to Qwen3-14B gains (+5.09/+15.32), confirming practitioners can select the backbone that fits their compute budget.
For domain specialization, Figure~\ref{fig:ablation}(b) shows even 200 samples yield gains over the no-DA baseline, with diminishing returns beyond 500, making it practical for deployment where large-scale annotation is costly.

\section{Related Work}\label{sec:related_works}
\noindent\textbf{LLM-as-a-Judge.}
As LLM is becoming increasingly powerful, more LLM-related applications are emerging~\cite{xue2026inference,ma2026autodata,ma2026non,song2025learning,ma2025llm}. LLM-as-a-Judge, employing LLMs as automated evaluators~\cite{zheng2023judging, kim2024prometheus}, has been widely used recently in these applications~\cite{qian2026brick,li2025verification}. However, general-purpose LLMs suffer from systematic biases~\cite{chen2024humans, wang2024llmjudgebias}, motivating dedicated judge training.
SFT-based approaches such as Self-Taught Evaluators~\cite{wang2024selftaught}, Skywork-Critic~\cite{liu2026skywork}, ArmoRM~\cite{wang2024interpretable}, and DeepSeek-GRM~\cite{liu2025inference} train judges through synthetic data curation, but rely on pattern memorization rather than flexible reasoning~\cite{chen2025judgelrm}.
Among RL-based approaches, JudgeLRM~\cite{chen2025judgelrm} demonstrates that judgment is inherently reasoning-intensive;
Think-J~\cite{huang2026thinkj} proposes judgment thinking initialization followed by GRPO-based optimization;
CompassJudger-2~\cite{zhang2025compassjudger2} combines task-driven data curation with verifiable rewards;
and J1~\cite{whitehouse2026j1} introduces consistency-based rewards both response orders, though neither code nor checkpoints have been publicly released.
CollabEval~\cite{qian2026collabeval} demonstrates that multi-agent collaboration improves evaluation quality but requires deploying multiple large models at inference time. Inspired by this, we aim to train a compact model that internalizes multi-agent deliberation capability via adaptive multi-reward RL, eliminating the cost of multi-model deployment.

\noindent\textbf{Multi-Reward RL Optimization.}
Recent LLM training increasingly incorporates multiple reward signals, such as correctness + format compliance~\cite{chen2025judgelrm}.
GDPO~\cite{liu2026gdpo} identifies that GRPO suffers from reward signal collapse in multi-reward settings and proposes decoupled per-reward normalization.
Lu et al.~\cite{lu2026learning} propose dynamic reward weighting using hypervolume-guided adaptation and gradient-based weight optimization to explore Pareto fronts.
However, these approaches either use fixed weights~\cite{aggarwal2025l1, chen2025judgelrm}, require manual reward function conditioning~\cite{liu2026gdpo}, or need expensive per-objective gradient computation~\cite{lu2026learning}.
Our AdaReward uses a lightweight learner that requires only scalar reward statistics, making it computationally cheap while achieving adaptive behavior.

\section{Conclusion}

In this work, we propose \modelname, a novel framework that equips a single compact judge model with multi-agent panel deliberation capability via adaptive multi-reward RL.
Through extensive experiments on four established judge benchmarks, we demonstrate that JudgePanel (14B) consistently outperforms judge-specialized models (up to 70B), while maintaining strong position consistency.
Furthermore, it also enables rapid specialization to new evaluation tasks with few hundred samples, demonstrating its practicality for real-world deployment.
We believe JudgePanel provides an effective and scalable solution for practitioners who need reliable, domain-specific evaluation capabilities without the overhead of deploying large-scale models or multi-agent systems in production.
\section*{Limitations}\label{appendix: limitations}

\textbf{Inference Token Cost.}
Due to the multi-agent reasoning trace, JudgePanel generates more tokens per evaluation than standard COT-based judges.
The panel structure (independent evaluation round, discussion rounds, and final judgment) inherently requires longer outputs to simulate multi-speaker deliberation.
However, as shown in Table~\ref{tab:analysis}(b), JudgePanel still achieves competitive latency (1.5 min vs.\ 1.0 min for Think-J on 100 samples) while outperforming COT-based Think-J in accuracy (91.33 vs.\ 90.50 on RewardBench, 45.61 vs.\ 40.15 on PPE) with a smaller model (14B vs.\ 32B).
The accuracy gains justify the additional token cost for applications where judgment quality is prioritized over throughput.

\noindent\textbf{Domain Specialization Pipeline.}
While JudgePanel can quickly adapt to new domains with as few as 200 labeled samples, the current domain specialization pipeline still requires running the panel protocol on these labeled samples to generate reasoning traces for LoRA fine-tuning.
This introduces a dependency on the panel models during the specialization phase.
Exploring more efficient domain specialization strategies, such as directly fine-tuning on verdict labels without trace generation, applying few-shot learning to the trained JudgePanel, or using the trained JudgePanel itself to generate specialization traces, is a promising direction for future research.
\clearpage
\bibliography{custom}

\clearpage
\appendix
\section*{Appendix}

This appendix provides supplementary material for the main paper.
\Aref{appendix:reward_details} details the reward function scoring.
\Aref{appendix:algorithm} lists the full training framework.
\Aref{appendix:experiment_details} presents experiment details including hyperparameters, benchmarks, and evaluation metrics.
\Aref{appendix:case_examples} provides qualitative case examples.

\section{Reward Function Details}\label{appendix:reward_details}

We design $K=4$ reward functions that jointly capture both judgment correctness and reasoning process quality. 
Unlike existing approaches that use a single binary outcome reward~\cite{whitehouse2026j1, zhang2025compassjudger2}, our multi-aspect design provides fine-grained learning signals: $R_{\text{out}}$ diagnoses where errors originate in the reasoning chain; $R_{\text{dis}}$ ensures genuine deliberation independent of outcome; $R_{\text{eff}}$ penalizes counterproductive discussion that corrupts correct initial judgments; and $R_{\text{fmt}}$ enforces the structural format necessary for the other rewards to be computed. Together, these rewards create complementary training signals that cannot be captured by any single reward function.

\noindent\textbf{Fine-Grained Outcome Reward} $R_{\text{out}}$.
Rather than a binary correct/incorrect signal, this reward diagnoses \textit{where} the reasoning went wrong: whether the error originated in the initial evaluation, was introduced during discussion, or occurred at the final judgment stage. 
This provides partial credit when the panel was on the right track.
Let $\hat{y}_i$ denote the final verdict, $y_i$ the ground truth, $\vv_{\text{init}} = \{v_i^{(1)}, \ldots, v_i^{(M)}\}$ the initial speaker verdicts, and $\vv_{\text{last}}$ the verdicts from the last stage before the final result. $\text{maj}(\cdot)$ denotes the majority verdict.
\begin{equation}
    R_{\text{out}}(\tau_{i,j}) {=} \begin{cases}
        1.0 & \hat{y}_i {=} y_i, \\
        0.7 & \hat{y}_i {\neq} y_i,\, \text{maj}(\vv_{\text{last}}) {=} y_i, \\
        0.5 & \hat{y}_i {\neq} y_i,\, \text{maj}(\vv_{\text{init}}) {=} y_i, \\
        0.3 & \hat{y}_i {\neq} y_i,\, y_i {\in} \vv_{\text{init}}, \\
        0.0 & \text{otherwise}.
    \end{cases}
\end{equation}

\noindent\textbf{Discussion Quality Reward} $R_{\text{dis}}$.
This reward measures whether the multi-round discussion produces genuine deliberation that contributes to correct outcomes.
Let $d$ denote the number of discussion rounds and $\Delta_{\text{score}}$ indicate whether any speaker changed their verdict:
\begin{equation}
  r_{\mathrm{disc}} =
  \begin{cases}
  1.0 & d=0,\ \text{unanimous},\ \hat{y}=y,\\
  0.0 & d=0,\ \text{unanimous},\ \hat{y}\neq y,\\
  0.2 & d=0,\ \lnot\,\text{unanimous (initial split)},\\
  0.2 & d>0,\ \mathbf{v}^{(d)}\ \text{unparseable},\\
  1.0 & d>0,\ \Delta_{\mathrm{score}},\ \hat{y}=y,\\
  0.3 & d>0,\ \Delta_{\mathrm{score}},\ \hat{y}\neq y,\\
  0.3 & d>0,\ \lnot\,\Delta_{\mathrm{score}}.
  \end{cases}
  \label{eq:discussion_reward}
  \end{equation}

 Intuitively, the two extreme signals are the unanimous no-discussion cases: a
  panel that already agrees and is right ($1.0$) needs no deliberation, while a
  confidently-wrong consensus ($0.0$) is maximally penalized. When the panel is
  split but skips discussion ($d=0$, not unanimous), it receives a low $0.2$ for
  failing to deliberate. Once discussion occurs ($d>0$), the panel earns the
  full $1.0$ only when scores actually change and the final verdict is
  correct, rewarding productive revision; a discussion that changes scores
  but stays wrong, or that merely restates positions without any change, both
  receive $0.3$. The residual $0.2$ covers rounds whose scores cannot be parsed.
  
\noindent\textbf{Efficiency Reward} $R_{\text{eff}}$.
This reward evaluates whether the model's decision to engage in or skip discussion was \textit{beneficial} to the final outcome. The strongest penalty ($-0.3$) targets the most harmful failure: when initial speakers had the correct majority but discussion corrupted it, directly discouraging counterproductive deliberation.
\begin{equation}
    R_{\text{eff}}(\tau_{i,j}) = \begin{cases}
        1.0 & \hat{y}_i {=} y_i,\, \text{unanimous},\, d{=}0, \\
        0.8 & \hat{y}_i {=} y_i,\, \text{discussion corrected}, \\
        0.3 & \hat{y}_i {=} y_i,\, \text{otherwise}, \\
        -0.3 & \hat{y}_i {\neq} y_i,\, \text{flipped correct}, \\
        0.0 & \hat{y}_i {\neq} y_i,\, \text{otherwise}.
    \end{cases}
\end{equation}

\noindent\textbf{Format Compliance Reward} $R_{\text{fmt}}$.
Computed as a weighted sum of binary checks for:
(i) required section tags (e.g., \texttt{[Initial Evaluation]}, \texttt{[Final Evaluation Result]}),
(ii) presence of $M$ speakers with complete evaluation fields (strengths/weaknesses, per-criterion scores, justification, confidence), and
(iii) proper discussion and final judge structure when present.

\section{Algorithm}\label{appendix:algorithm}

Algorithm~\ref{alg:pipeline} summarizes the complete \modelname training pipeline, including panel data construction, SFT, AdaReward reinforcement learning with adaptive weight updates, and lightweight domain specialization.

\begin{algorithm}[H]
\caption{\modelname: Training Pipeline}
\label{alg:pipeline}
\begin{algorithmic}[1]
\renewcommand{\algorithmicrequire}{\textbf{Input:}}
\renewcommand{\algorithmicensure}{\textbf{Output:}}
\Require Base model $\pi_{\theta_0}$; panel model $\gP$; arena data $\gD_{\text{arena}}$; rewards $\{R_k\}_{k=1}^{K}$; targets $\{\rho_k\}$
\Ensure Domain-specialized judge $\pi_\theta$
\State \textbf{Stage 1: Panel Data Construction} (\S\ref{sec:data})
\State Run panel protocol on $\gD_{\text{arena}}$; retain correct traces $\to \gD_{\text{panel}}$
\State \textbf{Stage 2: SFT}
\State Fine-tune $\pi_\theta$ on $\gD_{\text{panel}}$
\State \textbf{Stage 3: AdaReward RL} (\S\ref{sec:adgdpo})
\For{step $s = 1, \ldots, S$}
    \State Sample rollouts, compute rewards $\{R_k(\tau_{i,j})\}$
    \State Compute $\hat{A}_{\text{sum}}^{(i,j,s)}$ via adaptive weights 
    \State Update $\pi_\theta$ via $\gJ_{\text{Ada}}$ (Eq.~\ref{eq:adgdpo})
    \If{$s \mod \Delta = 0$}
        \State Update weights viaEq.~\ref{eq:target_weight}--\ref{eq:ema}
    \EndIf
\EndFor
\State \textbf{Stage 4: Domain Specialization} (\S\ref{sec:da})
\State LoRA fine-tune on domain-specific panel traces (200--500 samples)
\State
\Return $\pi_\theta$
\end{algorithmic}
\end{algorithm}

\section{Experiment Details}\label{appendix:experiment_details}

This section provides implementation details for reproducing our results, including training hyperparameters for all three stages, benchmark descriptions, domain specialization data sources, and evaluation metric definitions.

\subsection{Hyper-Parameters}\label{appendix:hyperparameters}
For SFT, we use LlamaFactory \footnote{\url{http://github.com/hiyouga/LlamaFactory}}~\cite{zheng2024llamafactory} train for 3 epochs with learning rate $1 \times 10^{-6}$, patience 10, cosine schedule, batch size 32, and max sequence length 8,192.
For AdaReward RL, we train for 5 epochs with $G{=}8$ rollouts per prompt, learning rate $5 \times 10^{-7}$, clip ratio $\varepsilon{=}0.2$, and KL penalty $0.08$.
The adaptive learner uses: initial weights $w_{\text{out}}^{(0)}{=} 5.0$, $w_{\text{disc}}^{(0)}{=} 2.0$, $w_{\text{fmt}}^{(0)}{=} 2.0$, $w_{\text{eff}}^{(0)}{=} 2.0$, targets $\rho_{\text{out}}{=}0.95$, $\rho_{\text{dis}}{=}0.85$, $\rho_{\text{eff}}{=}0.85$, $\rho_{\text{fmt}}{=}0.99$, scaling $\alpha^+{=}0.5$, $\alpha^-{=}0.1$, momentum $\beta{=}0.3$, bounds $[3, 10]$ for outcome and $[1, 4]$ for other rewards, and update interval $\Delta{=}5$.
For DA, we apply LoRA ($r{=}16$, $\alpha{=}32$) with learning rate $1 \times 10^{-4}$ and early stop with patient 10 for 3 epochs on  domain-specific samples.
All experiments are conducted on 8$\times$ NVIDIA H200 (143GB) GPUs using EasyR1\footnote{\url{https://github.com/hiyouga/EasyR1}}~\cite{zheng2025easyr1} for the implementation of GDPO and GRPO. SFT training takes approximately 2 hours, RL training takes approximately 22 hours, and LoRA DA takes approximately 5 minutes per task.

\subsection{Benchmark and DA Data Details}\label{appendix:benchmarks}

\textbf{Benchmarks.}
(1)~JudgeBench~\cite{tan2025judgebench}: 620 challenging response pairs spanning knowledge, reasoning, math, and coding, evaluated by objective correctness.
(2)~RewardBench~\cite{lambert2025rewardbench}: 2,985 prompt-chosen-rejected trios covering chat, reasoning, and safety.
(3)~RMBench~\cite{liu2025rmbench}: 3,981 samples testing reward model sensitivity to subtle content differences and resistance to style biases.
(4)~PPE~\cite{frick2025evaluate}: 16,038 samples evaluating pairwise preference alignment across 12 domains with multilingual coverage.
For fair comparison, all models use \textit{parsed accuracy}, computed as correct predictions divided by total samples excluding unparseable outputs.

\subsection{Evaluation Metric Definitions}\label{appendix:metrics}

\textbf{Avg.\ Rank} (Table 1): For each benchmark, models are ranked by accuracy (rank 1 = highest). Avg.\ Rank is the mean rank across all four benchmarks (lower is better).

\noindent\textbf{Consistency} (Table 2): The percentage of samples where the model's verdict correctly tracks position swaps, i.e., if the model prefers Response A in the original order, it should prefer Response B when the order is swapped.

\noindent\textbf{Stable Accuracy} (Table 2): The percentage of samples correctly predicted in \textit{both} orderings, measuring position-invariant judgment quality.

\noindent\textbf{Constructive Change Rate} (Table 3a): Among samples where at least one speaker changes their vote during discussion, the percentage that end up with the correct final answer. Formally: $|\{\text{vote changed} \cap \text{final correct}\}| / |\{\text{vote changed}\}|$. A rate near 50\% indicates random changes; significantly above 50\% indicates deliberation is functional.

\noindent\textbf{Jaccard Distance} (Table 3a, speaker-level, $\uparrow$): Computed as $1 - |T_i \cap T_j| / |T_i \cup T_j|$ over character 3-grams between each pair of speakers within the same round, then averaged across all pairs. Higher values indicate speakers use more distinct phrasing and reasoning paths.

\noindent\textbf{4-gram Overlap} (Table 3a, phase-level, $\downarrow$): The fraction of word 4-grams in discussion rounds that also appear in the initial round: $|N_{\text{disc}} \cap N_{\text{init}}| / |N_{\text{disc}}|$. Lower values indicate discussion introduces novel content rather than repeating prior reasoning.

\noindent\textbf{Average Rounds} (Table 3a): The mean number of discussion rounds generated per sample. The model should adaptively engage in discussion only when speakers initially disagree.

\section{Case Examples}\label{appendix:case_examples}

We provide qualitative examples comparing panel trace generation with single-model COT generation on the same evaluation task, illustrating how multi-agent deliberation enables disagreement resolution that single-model reasoning cannot achieve.

\subsection{Single COT Think Trace}\label{appendix:single_cot_trace}

Below is a single-model COT thinking trace generated by Qwen model on the same coding evaluation task, following the format of independent evaluation of a single speaker. Without multi-agent discussion, the model arrives at an incorrect verdict.

\begin{tcolorbox}[colback=gray!5!white,colframe=gray!75!black, title={Input}, boxsep=1mm, top=1mm, bottom=1mm]
\footnotesize
\textbf{Instruction:} You are given two integers \texttt{num1} and \texttt{num2}. In one operation, you can choose integer $i$ in the range $[0, 60]$ and subtract $2^i + \texttt{num2}$ from \texttt{num1}. Return the minimum number of operations needed to make \texttt{num1} equal to 0. If impossible, return $-1$.

\textbf{Response A:}
\begin{verbatim}
for k in range(61):
  target = num1 - k * num2
  if target >= 0 and bin(target).count('1') <= k:
    return k
return -1
\end{verbatim}

\textbf{Response B:}
\begin{verbatim}
for k in range(61):
  target = num1 - k * num2
  if target < 0: continue
  if bin(target).count('1') <= k and k <= target:
    return k
return -1
\end{verbatim}
\end{tcolorbox}

\begin{tcolorbox}[colback=myblue!5!white,colframe=myblue!75!black, title={Single COT Evaluation}, boxsep=1mm, top=1mm, bottom=1mm, breakable]
\footnotesize
Let me analyze both solutions. After $k$ operations, we need $\texttt{num1} - k \cdot \texttt{num2} = \Sigma(2^{i_j})$. The minimum powers of two needed equals the popcount. Both solutions check this condition. Response B adds \texttt{k <= target} and explicit negative-target skip. However, \texttt{k <= target} appears redundant within problem constraints ($\texttt{num1} \leq 10^9$): when \texttt{num2} is negative, target grows with $k$; when positive, the loop terminates early. Response A captures the essential logic without unnecessary conditions.

\textbf{Strengths and Weaknesses of Response A:} Elegant and minimal. Correctly identifies the two core conditions (non-negative target and bit-count bound). No unnecessary imports or redundant checks.

\textbf{Strengths and Weaknesses of Response B:} Functionally correct but over-engineered. The \texttt{k <= target} condition is redundant within problem constraints, and the unused \texttt{math} import indicates incomplete code cleanup.

\textbf{Score of Response A:} Total: 42. Accuracy: 9, Relevance: 9, Completeness: 8, Clarity: 9, Usefulness: 7.

\textbf{Score of Response B:} Total: 36. Accuracy: 8, Relevance: 9, Completeness: 7, Clarity: 7, Usefulness: 5.

\textbf{Potential Errors:} Edge cases where \texttt{num2} is positive and large could make target small relative to $k$, but the \texttt{target >= 0} guard handles termination.

\textbf{Evaluation Result:} \underline{model\_a}

\textbf{Justification:} Response A is preferred for its clean, minimal implementation that captures exactly the necessary and sufficient conditions.
\end{tcolorbox}

\subsection{Panel Trace Example}\label{appendix:panel_trace_full}

Below is a full panel reasoning trace generated by our multi-agent protocol on the same coding evaluation task. In contrast to the single COT above, the panel protocol enables disagreement resolution: Speaker 2 initially favors model\_a (similar to the single COT reasoning), but after hearing Speaker 1's argument about Python's \texttt{bin()} behavior on negatives during discussion, revises to model\_b. All three speakers then reach consensus with maximum confidence---demonstrating how multi-agent deliberation corrects individual reasoning errors.

\begin{tcolorbox}[colback=gray!5!white,colframe=gray!75!black, title={Input}, boxsep=1mm, top=1mm, bottom=1mm, breakable]
\footnotesize
\textbf{Instruction:} You are given two integers \texttt{num1} and \texttt{num2}. In one operation, you can choose integer $i$ in the range $[0, 60]$ and subtract $2^i + \texttt{num2}$ from \texttt{num1}. Return the minimum number of operations needed to make \texttt{num1} equal to 0. If impossible, return $-1$.

\textbf{Response A:}
\begin{verbatim}
for k in range(61):
  target = num1 - k * num2
  if target >= 0 and bin(target).count('1') <= k:
    return k
return -1
\end{verbatim}

\textbf{Response B:}
\begin{verbatim}
for k in range(61):
  target = num1 - k * num2
  if target < 0: continue
  if bin(target).count('1') <= k and k <= target:
    return k
return -1
\end{verbatim}
\end{tcolorbox}

\begin{tcolorbox}[colback=myblue!5!white,colframe=myblue!75!black, title={Phase 1: Independent Evaluation}, boxsep=1mm, top=1mm, bottom=1mm, breakable]
\footnotesize
\textbf{[Speaker 1]} Score: model\_b | Confidence: 5

After $k$ operations, $\Sigma(2^{i_j}) = \texttt{num1} - k \cdot \texttt{num2}$. Each term $2^{i_j} \geq 1$, so target $\geq k$. Response A has a bug: Python's \texttt{bin()} on negatives returns \texttt{'-0b...'} where \texttt{count('1')} yields 0, falsely satisfying the condition. Response B explicitly skips negatives and adds safe $k \leq$ target bound.

\textbf{Strengths and Weaknesses of Response A:} Concise and clean but lacks proper handling of negative targets---Python's \texttt{bin()} on negative numbers produces unexpected output, leading to incorrect acceptance of invalid cases.

\textbf{Strengths and Weaknesses of Response B:} More robust by explicitly skipping negative targets. The added \texttt{k <= target} condition is technically redundant but harmless.

\textbf{Score of Response A:} Total: 33. Accuracy: 6, Relevance: 8, Completeness: 7, Clarity: 8, Usefulness: 4.

\textbf{Score of Response B:} Total: 40. Accuracy: 9, Relevance: 8, Completeness: 8, Clarity: 8, Usefulness: 7.

\textbf{Potential Errors:} One might overlook the subtle issue with Python's \texttt{bin()} behavior on negative numbers.

\textbf{Evaluation Result:} model\_b

\textbf{Justification:} Response B avoids the Python-specific bug where \texttt{bin()} on negative numbers produces malformed strings that incorrectly satisfy the bit-count condition.

\rule{\linewidth}{0.4pt}

\textbf{[Speaker 2]} Score: model\_a | Confidence: 5

Both solutions use the same core algorithm. Response A is cleaner and more concise. The \texttt{k <= target} condition in Response B appears redundant---since \texttt{bin(target).count('1') <= k} already implies the target has enough bits to decompose into $k$ powers. Response B also has an unnecessary \texttt{import math}. The extra condition adds complexity without clear benefit.

\textbf{Strengths and Weaknesses of Response A:} Clean, minimal implementation. Directly checks the two necessary conditions without unnecessary additions. Easy to read and maintain.

\textbf{Strengths and Weaknesses of Response B:} Adds a seemingly redundant \texttt{k <= target} check and an unused import. While not incorrect, the extra complexity is unjustified and suggests the author may not fully understand why the condition is unnecessary.

\textbf{Score of Response A:} Total: 40. Accuracy: 8, Relevance: 9, Completeness: 8, Clarity: 9, Usefulness: 6.

\textbf{Score of Response B:} Total: 36. Accuracy: 7, Relevance: 9, Completeness: 7, Clarity: 7, Usefulness: 6.

\textbf{Potential Errors:} May be underestimating edge cases where the redundant condition actually matters.

\textbf{Evaluation Result:} model\_a

\textbf{Justification:} Response A is preferred for its simplicity and directness. The additional checks in B appear unnecessary and add complexity without improving correctness.

\rule{\linewidth}{0.4pt}

\textbf{[Speaker 3]} Score: model\_b | Confidence: 4

Response A lacks an explicit check for the case where $k$ exceeds the target, which could lead to incorrect results. Response B improves upon A by including this additional check.

\textbf{Strengths and Weaknesses of Response A:} Straightforward approach but lacks explicit check for $k >$ target scenario.

\textbf{Strengths and Weaknesses of Response B:} More comprehensive and accurate by including the additional $k \leq$ target check.

\textbf{Score of Response A:} Total: 36. Accuracy: 8, Relevance: 9, Completeness: 6, Clarity: 8, Usefulness: 5.

\textbf{Score of Response B:} Total: 42. Accuracy: 9, Relevance: 9, Completeness: 8, Clarity: 8, Usefulness: 8.

\textbf{Potential Errors:} Not considering the scenario where $k$ exceeds the target could lead to incorrect conclusions.

\textbf{Evaluation Result:} model\_b

\textbf{Justification:} Response B is better due to its improved completeness and accuracy from the additional check preventing $k >$ target errors.

\vspace{1mm}
\textit{\textcolor{red}{$\rightarrow$ No consensus (S1, S3: model\_b; S2: model\_a). Proceeding to discussion.}}
\end{tcolorbox}

\begin{tcolorbox}[colback=myred!5!white,colframe=myred!75!black, title={Phase 2: Discussion (Round 1)}, boxsep=1mm, top=1mm, bottom=1mm, breakable]
\footnotesize
\textbf{[Speaker 1]} Revised Score: model\_b | Confidence: 5

\textbf{Disagreement Resolution:} Speaker 2 argues that \texttt{k <= target} is redundant and prefers A for conciseness. However, this overlooks the critical Python-specific bug: \texttt{bin()} on negative numbers returns \texttt{'-0b101'} whose \texttt{.count('1')} yields 0, falsely satisfying the condition. Response A's \texttt{target >= 0} check does guard against this, but combined with the missing $k \leq$ target bound, A has two potential failure modes.

\textbf{Revised Score of Response A:} Total: 34. Accuracy: 6, Relevance: 8, Completeness: 7, Clarity: 8, Usefulness: 5.

\textbf{Revised Score of Response B:} Total: 41. Accuracy: 9, Relevance: 9, Completeness: 9, Clarity: 8, Usefulness: 8.

\textbf{Justification:} Response B's explicit handling of negative targets and adherence to all logical requirements outweigh conciseness concerns raised by Speaker 2.

\rule{\linewidth}{0.4pt}

\textbf{[Speaker 2]} Revised Score: model\_b | Confidence: 5

\textbf{Disagreement Resolution:} I originally overweighted conciseness and claimed \texttt{k <= target} was redundant. \underline{Speaker 1 correctly identified a deeper issue I missed}: Python's \texttt{bin()} on negative numbers returns strings like \texttt{-0b101} whose \texttt{.count('1')} yields 0, falsely satisfying the condition. While Response A does check \texttt{target >= 0}, the combination of this subtle behavior with the missing $k \leq$ target bound creates real correctness vulnerabilities. \underline{I am revising my position from model\_a to model\_b}.

\textbf{Revised Score of Response A:} Total: 33. Accuracy: 5, Relevance: 8, Completeness: 7, Clarity: 8, Usefulness: 5.

\textbf{Revised Score of Response B:} Total: 43. Accuracy: 9, Relevance: 9, Completeness: 9, Clarity: 8, Usefulness: 8.

\textbf{Justification:} Speakers 1 and 3 correctly identified that Response B covers all bases safely. Response A's absence of an explicit \texttt{target < 0} guard and missing $k \leq$ target creates correctness vulnerabilities that outweigh its conciseness advantage.

\rule{\linewidth}{0.4pt}

\textbf{[Speaker 3]} Revised Score: model\_b | Confidence: 5

\textbf{Disagreement Resolution:} Speaker 2 initially preferred Response A for conciseness but has now revised after acknowledging the \texttt{bin()} pitfall identified by Speaker 1. I concur---the omitted negative-target guard in A is a concrete bug. Raising confidence from 4 to 5 given full consensus.

\textbf{Revised Score of Response A:} Total: 33. Accuracy: 5, Relevance: 9, Completeness: 6, Clarity: 8, Usefulness: 5.

\textbf{Revised Score of Response B:} Total: 44. Accuracy: 9, Relevance: 9, Completeness: 9, Clarity: 8, Usefulness: 9.

\textbf{Justification:} All speakers recognized that Response B handles both the Python-specific pitfall and the required \texttt{k <= target} bound, making it the only solution guaranteed to avoid false positives.

\vspace{1mm}
\textit{\textcolor{red}{$\rightarrow$ Consensus reached. All speakers agree on model\_b with confidence 5. Final evaluation result:}}
\end{tcolorbox}

\begin{tcolorbox}[colback=green!5!white,colframe=green!50!black, title={Final Evaluation Result}, boxsep=1mm, top=1mm, bottom=1mm]
\footnotesize
\textbf{Score: model\_b | Confidence: 5}
\end{tcolorbox}


\newpage

\end{document}